\documentclass[10pt,twocolumn,letterpaper]{article}

\usepackage{iccv}
\usepackage{graphicx}
\usepackage{amsmath}
\usepackage{amssymb}
\usepackage{booktabs}
\usepackage{multirow}
\usepackage[noend]{algpseudocode}
\usepackage{algorithm}
\usepackage{siunitx}
\usepackage{amssymb}
\usepackage{pifont}
\usepackage{subcaption,float}

\usepackage[pagebackref,breaklinks,colorlinks]{hyperref}

\newcommand{\cmark}{\ding{51}}%
\newcommand{\xmark}{\ding{55}}%

\usepackage[capitalize]{cleveref}
\crefname{section}{Sec.}{Secs.}
\Crefname{section}{Section}{Sections}
\Crefname{table}{Table}{Tables}
\crefname{table}{Tab.}{Tabs.}

\newcommand*\matr{\mathbf}
\newcommand*\trans{\text{T}}

\newcommand{\abest}[1]{\underline{\textbf{#1}}}

\newcommand{\Hom}{\matr{H}}
\newcommand{\Aff}{\matr{A}}
\newcommand{\Fund}{\matr{F}}
\newcommand{\Ess}{\matr{E}}
\newcommand{\Intrinsic}{\matr{K}}
\newcommand*\Point{\mathbf{p}}
\newcommand*\Qoint{\mathbf{q}}
\newcommand{\Rot}{\matr{R}}
\newcommand{\Tran}{\mathbf{t}}

\newcommand{\Normline}{\mathbf{n}}
\newcommand{\Line}{\mathbf{l}}
\newcommand{\RR}{\mathbb{R}}
\newcommand{\Vertical}{\mathbf{v}}

\makeatletter
\def\BState{\State\hskip-\ALG@thistlm}
\makeatother

\algnewcommand\algorithmicforeach{\textbf{for each}}
\algdef{S}[FOR]{ForEach}[1]{\algorithmicforeach\ #1\ \algorithmicdo}

\algrenewcommand\algorithmicrequire{\textbf{Input:\phantom{ll}}}
\algrenewcommand\algorithmicensure{\textbf{Output:}}

\let\emptyset\varnothing

\newcommand*\Points{\mathcal{P}}

\iccvfinalcopy 

\def\iccvPaperID{7587} 

\ificcvfinal\fi

\begin{document}

\title{AffineGlue: Joint Matching and Robust Estimation}

\author{Daniel Barath$^1$, Dmytro Mishkin$^{2,4}$, \\ Luca Cavalli$^1$, Paul-Edouard Sarlin$^1$, Petr Hruby$^1$, Marc Pollefeys$^{1,3}$\\
$^1$ \normalsize{Department of Computer Science, ETH Zurich}, 
$^2$ \normalsize{Visual Recognition Group, CTU in Prague},\\
$^3$ \normalsize{Microsoft Mixed Reality and AI Zurich Lab},
$^4$ \normalsize{HOVER Inc.}
\\
{\tt\small dbarath@ethz.ch}
}

\ificcvfinal\thispagestyle{empty}\fi

\twocolumn[{
\maketitle
\vspace{-1em}
    \centering
    \scriptsize
    \setlength{\tabcolsep}{8pt}
    \newcommand{\sz}{0.90}
    \includegraphics[width=\sz\textwidth]{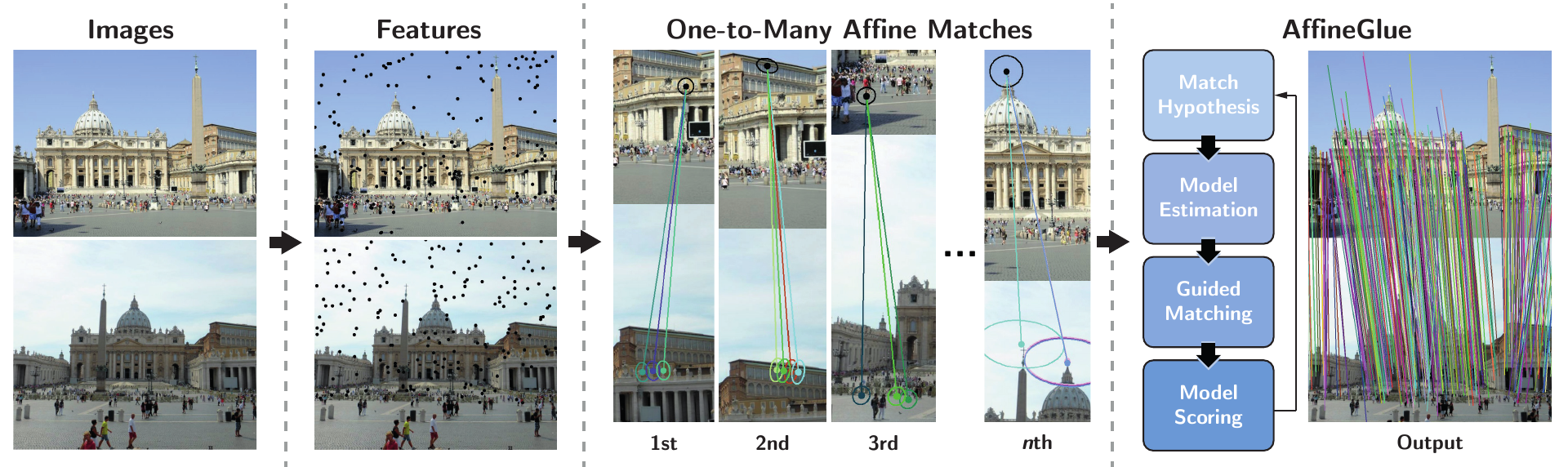}
    \captionof{figure}{The steps of the \textbf{AffineGlue pipeline} are as follows: 
    (1) features with affine shapes are detected in the input images, \eg, by SuperPoint~\cite{SuperPoint2017} combined with AffNet~\cite{AffNet2018}. 
    (2) For each feature in the source image, the matching by, \eg SuperGlue~\cite{sarlin2020superglue}, is often ambiguous, especially, at repeated patterns.
    Thus, we form \textit{one-to-many} matches for each point in the source image.
    (3) AffineGlue iteratively selects a candidate one-to-one affine correspondence and estimates the model (\eg, relative pose) by a single-point solver.
    Guided sampling then forms one-to-one correspondences
    consistent with the estimated model to calculate its score and select its inliers. }
    \label{fig:teaser}
\vspace{6pt}
}]

\maketitle

\begin{abstract}\vspace{-8pt}
We propose AffineGlue, a method for joint two-view feature matching and robust estimation that reduces the combinatorial complexity of the problem by employing single-point minimal solvers.
AffineGlue selects potential matches from one-to-many correspondences to estimate minimal models.
Guided matching is then used to find matches consistent with the model, suffering less from the ambiguities of one-to-one matches.
Moreover, we derive a new minimal solver for homography estimation, requiring only a single affine correspondence (AC) and a gravity prior.
Furthermore, we train a neural network to reject ACs that are unlikely to lead to a good model.
AffineGlue is superior to the SOTA on real-world datasets, even when assuming that the gravity direction points downwards. 
On PhotoTourism, the AUC@10$^\circ$ score is improved by 6.6 points compared to the SOTA. 
On ScanNet, AffineGlue makes SuperPoint and SuperGlue achieve similar accuracy as the detector-free LoFTR.
\end{abstract}

\section{Introduction}
\label{sec:intro}

Matching two or more images of a scene is a fundamental problem in vision with a wide range of applications, such as image retrieval~\cite{SIFT2004, arandjelovic2016netvlad, radenovic2016cnn, tolias2016image, noh2017large}, 
structure-from-motion~\cite{agarwal2011building, jared2015reconstructing,schonberger2016structure,zhu2018very, barath2021efficient}, 
localization~\cite{sattler2012improving, sattler2018benchmarking, lynen2020large, panek2022meshloc}, 
SLAM~\cite{engel2014lsd, mur2015orb, detone2017toward, detone2018superpoint}, 
and multi-view stereo~\cite{furukawa2010towards, furukawa2015multi, kar2017learning, Chen_2019_ICCV}.  
The traditional image matching pipeline consists of three main steps -- local feature detection, matching, and geometric robust estimation.
Due to this consecutive nature, matching failures often lead to failure in subsequent geometric robust estimation, rendering the pipeline unreliable.
While recent algorithms~\cite{sun2021loftr,wang2022matchformer,chen2022aspanformer} perform feature detection and matching jointly, at the cost of significantly increased run-time for all-pair 3D reconstruction, there is still a gap in the literature for methods that allow for simultaneous matching and robust estimation.
To address this issue, we propose a novel approach called \textit{AffineGlue} that employs joint feature matching and robust estimation by iteratively selecting potential matches, estimating the model, and performing guided matching to calculate the model score, \eg, via its support. 
While most methods need to commit to one-to-one matches to keep the problem tractable, we relax this to one-to-$k$ matches.

\vspace{1mm}\noindent\textbf{Feature detection and matching.}
Local features have been and still are the main workhorse in 3D reconstruction. 
Traditionally, local features involve three main steps: (scale-covariant) keypoint detection, orientation estimation, and descriptor extraction. 
Keypoint detection is typically performed on the scale pyramid employing a handcrafted response function, such as Hessian~\cite{Hessian78,HesHarAff2004}, Harris~\cite{Harris88alvey,HesHarAff2004}, Difference of Gaussians (DoG~\cite{SIFT2004}), or learned ones like FAST~\cite{FAST2006} or Key.Net~\cite{KeyNet2019}. 
The keypoint detection provides a triplet $(x,y, \text{scale})$ that defines a square or circular patch. 
The patch orientation is then estimated using handcrafted methods like dominant gradient orientation~\cite{SIFT2004} or center of mass~\cite{ORB2011} or learned ones like~\cite{OriNet2016,AffNet2018,selfscaori2021}.
Optionally, the affine-covariant shape~\cite{Baumberg2000,AffNet2018} is estimated. 
Finally, the patch is geometrically rectified and described using local patch descriptors such as SIFT~\cite{SIFT2004}, HardNet~\cite{HardNet2017}, SOSNet~\cite{SoSNet2019}, and others.

Recent advances in deep learning have led to the development of feature detection and description methods that do not rely on patch extraction.
Methods like SuperPoint~\cite{SuperPoint2017}, R2D2~\cite{R2D22019}, D2Net~\cite{D2Net2019} and DISK~\cite{DISK2020} employ feedforward Convolutional Neural Networks (CNNs) and assume up-is-up image orientation. 
Some recent methods have proposed learning matching directly, such as SuperGlue~\cite{sarlin2020superglue}, while others skip the detection step entirely~\cite{sun2021loftr,wang2022matchformer,chen2022aspanformer}.

\vspace{1mm}\noindent\textbf{Robust Estimation.}
Feature matching often leads to a large number of outliers that are inconsistent with the scene geometry. 
This holds especially in wide-baseline cases, where the inlier ratio often falls below 10\%. 
Robust estimation is thus crucial to simultaneously find the sought model (\eg, relative pose) and the matches consistent with it (its inliers). 
Classical approaches employ a RANSAC-like~\cite{fischler1981random} hypothesize-and-verify strategy, iteratively applying minimal solvers~\cite{fischler1981random, hartley1997defense, hartley2003multiple, stewenius2006recent, kukelova2008automatic, kukelova2017clever} to random subsets of the input data until an all-inlier sample is found.
To improve upon RANSAC, various techniques have been developed, such as local optimization methods (LO-RANSAC, LO$^+$-RANSAC, and GC-RANSAC)~\cite{chum2003locally,lebeda2012fixing,barath2021graph}, 
advanced scoring functions (MLESAC, MSAC, MAGSAC, and MAGSAC++)~\cite{torr2000mlesac,barath2019magsac,barath2020magsac++,barath2022learning}, 
speed-ups using probabilistic sampling (PROSAC, NAPSAC, and P-NAPSAC)~\cite{chum2005matching,torr2002napsac,barath2020magsac++}, 
preemptive verification strategies (SPRT and SP-RANSAC)~\cite{chum2008optimal,barath2021space}, 
degeneracy checks (DEGENSAC, QDEGSAC, and NeFSAC)~\cite{chum2005two,frahm2006ransac,cavalli2022nefsac}, 
and methods for auto-tuning of the inlier threshold (MINPRAN and a contrario RANSAC)~\cite{stewart1995minpran,moisan2012automatic,riu2022classification}.

In recent years, several neural network-based algorithms have been proposed aiming at robust relative pose estimation.
Context normalization networks~\cite{cne2018} is the first paper on the topic proposing to use PointNet (MLP) with batch normalization~\cite{batchnorm2015} as a context mechanism. 
Attentive context normalization networks~\cite{acne2020} introduces a special architectural block for the task. 
Deep Fundamental matrix estimation~\cite{dfe2018} uses differentiable iteratively re-weighted least-squares with predicted weights. 
The OANet algorithm~\cite{oanet2019} introduces several architectural blocks for correspondence filtering.
Neural Guided RANSAC~\cite{brachmann2019neural} uses a CNe-like architecture with a different training objective.
A guided sampling algorithm exploits the predicted correspondence scores inside RANSAC to find accurate models early.
CLNet~\cite{clnet2021} introduces several algorithmic and architectural improvements  to remove gross outliers with iterative pruning. 
These techniques provide alternatives for tentative correspondence pre-filtering and weighting after the matches are formed.
A final least-squares fitting or RANSAC is then applied to obtain the model parameters from the kept matches.
 

\vspace{1mm}\noindent\textbf{Motivation and Contributions.}
Jointly performing feature matching and robust estimation is a problem of high complexity, making it impractical in the general case.
For example, when matching $n$ features in each image, the matching complexity is $n^2$. 
Injecting this into the complexity of robust estimation, we get $\binom{n^2}{m}$, where $m$ is the sample size to fit a minimal model, such as $m = 5$ for essential matrix estimation. This makes the probability of selecting an all-inlier sample that leads to a good model extremely low.
When having $1000$ features in each image and estimating an essential matrix, more than $10^{26}$ combinations must be tried. 

We propose a new method, \textit{AffineGlue}, to perform joint feature matching and robust estimation by employing single-point solvers~\cite{scaramuzza20111,barath2017theory,guan2019rotational, hajder2020relative,eichhardt2020relative, guan2021relative}.
This approach reduces the complexity of the joint procedure to that of the matching $\mathcal{O}(n^2)$, as $m = 1$ in this special case.
We use minimal solvers that estimate the two-view geometry from a single affine correspondence (AC) -- a feature that contains higher-order information about the underlying scene geometry~\cite{barath2017theory,barath2018efficient,eichhardt2020relative}.
Also, we propose a new one for estimating the homography from a single AC. 
\textit{AffineGlue} uses any off-the-shelf feature matcher to form one-to-many correspondences that are finalized when performing robust estimation and guided matching.
Additionally, we train a neural network~\cite{cavalli2022nefsac} to efficiently reject ACs likely to be inconsistent with the sought model.
The proposed method outperforms state-of-the-art feature detectors and matchers by a significant margin on a variety of real-world and large-scale datasets.

%


\section{Theoretical Background}

\noindent
\textbf{Affine correspondence} $(\Point_1, \Point_2, \Aff)$ is a triplet, where $\Point_1 = [u_1 \; v_1 \; 1]^\trans$ and $\Point_2 = [u_2 \; v_2 \; 1]^\trans$ are a corresponding homogeneous point pair in two images and $\Aff$
is a $2 \times 2$ linear transformation which is called \textit{local affine transformation}. 
For $\Aff$, we use the definition provided in~\cite{Molnar2014} as it is given as the first-order Taylor-approximation of the $\text{3D} \to \text{2D}$ projection function. 

\noindent
\textbf{Fundamental matrix} $(\Fund)$ is a $3 \times 3$ rank-$2$ matrix relating the corresponding points $\Point_1$, $\Point_2$ as:
\begin{equation}
    \Point_2^{\text{T}} \Fund \Point_1 = 0. \label{eq:fund_constraint}
\end{equation}

\noindent
\textbf{Essential matrix} $(\Ess \in \mathbb{R}^{3\times3})$ is related to $\Fund$ as
    $\Intrinsic'^{-\text{T}} \Ess \Intrinsic^{-1} = \Fund$,
%
where $\Intrinsic$, $\Intrinsic'$ are the intrinsic parameters of the cameras~\cite{hartley2003multiple}. 
\eqref{eq:fund_constraint} can be written as $\Point_2^{\text{T}} \Intrinsic'^{-\text{T}} \Ess \Intrinsic^{-1} \Point_1 = 0$. 
In the rest of the paper, we assume that the corresponding points $\Point_1$, $\Point_2$ have been premultiplied by matrices $\Intrinsic$, $\Intrinsic'$. 
This simplifies \eqref{eq:fund_constraint} to 
\begin{equation}
    \Point_2^{\text{T}} \Ess \Point_1 = 0.
    \label{eq:ess_constraint}
\end{equation}
Essential matrix $\Ess$ is decomposed as $\Ess = [\Tran]_{\times} \Rot$, where $\Rot \in \text{SO}(3)$, $\Tran \in \RR^3$ is the relative pose of the two views.

The relationship of an affine correspondence (AC) and essential matrix $\Ess$ was first defined in \cite{DBLP:journals/tip/BarathH18} as
\begin{equation}
    \Aff^{-\text{T}} \Normline_1 = -\Normline_2, \label{eq:aff_constraint}
\end{equation}
where $\Normline_1$, $\Normline_2$ are the normals to the epipolar lines in the images. 
This linear constraint is built on two properties of ACs. 
First, due to $\Aff$ being a linear approximation of the imaging function, it transforms the infinitesimal neighborhood of $\Point_1$ to that of $\Point_2$. 
Therefore, $\Aff$ maps the lines passing through $\Point_1$.
Thus, $\Aff \Point_1 \parallel \Point_2$ which can be written as
    $\Aff^{-\text{T}} \Normline_1 = \beta \Normline_2$,
where $\Normline_1$, $\Normline_2$ are the normals to the epipolar lines and operator $\parallel$ denotes two parallel vectors; $\beta \in \RR$.
These normals are calculated as the first two coordinates of the epipolar lines as $\Normline_1 = \Line_{1[1:2]} = (\Ess^{\text{T}} \Point_2)_{[1:2]}$, $\Normline_2 = \Line_{2[1:2]} = (\Ess \Point_1)_{[1:2]}$.
Since $\Normline_1$ and $\Normline_2$ absorb the scaling from $\Ess$, scalar $\beta$ is $-1$.
In summary, an affine correspondence imposes three independent constraints on the essential matrix.
One is given by \eqref{eq:ess_constraint}, and two others by \eqref{eq:aff_constraint}.

\section{Joint Matching and Estimation}

A method is proposed in this section to robustly estimates the parameters of the sought model while simultaneously performing feature matching. See Fig.~\ref{fig:teaser}.
The pseudo-code of the algorithm is as follows:
\begin{algorithmic}
\Require $\Points_1$, $\Points_2$ -- data points in the two images
\Ensure $\mathcal{M}^*$ -- correspondences, $\theta$ -- model params.\
\State $\theta^* \leftarrow \textbf{0}, q^* \leftarrow 0$, $\mathcal{M}^* \leftarrow \emptyset$ \Comment{Initialization}
\While{$\neg$Terminate()}
\State $\mathcal{S} \leftarrow \text{NextBestMatch}(\Points_1, \Points_2)$ \Comment{Generate a match}
\State $\theta \leftarrow \text{EstimateModel}(\mathcal{S})$ \Comment{A one-point solver}
\State $\mathcal{M} \leftarrow \text{GuidedMatching}(\theta, \Points_1, \Points_2)$ 
\State $q \leftarrow \text{GetScore}(\theta, \mathcal{M})$
\If{$q > q^*$} \Comment{Update the best model}
    \State $q', \theta', \mathcal{M}' \leftarrow \text{LocalOptimization}(\theta, \Points_1, \Points_2)$
    \State $\theta^* \leftarrow \theta', q^* \leftarrow q'$, $\mathcal{M}^* \leftarrow \mathcal{M}'$
\EndIf
\EndWhile
\end{algorithmic}
Similar to RANSAC, we formalize the problem as iterative sampling, model estimation, and scoring. 
We assume, however, to have a minimal solver that estimates the model parameters from a single match. 
This allows formalizing function \texttt{NextBestMatch} that forms sample $\mathcal{S}$ consisting of a single correspondence ($\textbf{p}_1$, $\textbf{p}_2$, $\Aff$) in each iteration, where $\textbf{p}_1 \in \Points_1$ and $\textbf{p}_2 \in \Points_2$ are points in the images, and $\Aff \in \mathbb{R}^{2\times2}$ is the local affine frame. 
Model $\theta$ is estimated from $\mathcal{S}$. 
Note that the method works with any single-point solver, \eg \cite{scaramuzza20111}, not only with ones leveraging ACs. 

After estimating the model, we perform guided matching~\cite{shah2015geometry,ma2018guided,barath2021efficient} using model $\theta$ to find a set $\mathcal{M}$ of correspondences consistent with the model parameters. 
The model quality $q$ is calculated from $\mathcal{M}$, \eg, as its support (\ie, $|\mathcal{M}|$), or by any existing scoring technique.
In case a new best model is found, we apply local optimization to improve its accuracy. 
The algorithm runs until the termination criterion is triggered. 
Next, we will describe each step in depth.

\vspace{1mm}\noindent\textbf{Next Best Match Selection.}
%
Suppose that we are given $n_1, n_2 \in \mathbb{N}^+$ features in the first and second images, respectively.
Forming correspondences is of quadratic complexity $\mathcal{O}(n_1 n_2)$. 
Thus, iterating through all possible matches, while doable, severely affects the run-time.
To alleviate this computational burden, we obtain the $k$ best matches for each feature in the source image, where $k \ll n_2$, $k \in \mathbb{N}^+$. 
This can be done by applying the standard $k$-nearest-neighbors ($k$NN) descriptor matching. 
Algorithms like SuperGlue, solving the optimal transport problem, provide a score matrix via the Sinkhorn algorithm~\cite{knight2008sinkhorn}.  
In this case, the $k$ best matches are the $k$ features with the highest scores.
%
This allows \textit{AffineGlue} to explore the $k$ best matches and thus, reduce the matching ambiguity -- for example, see Fig.~\ref{fig:teaser}, where the potential matches are on the windows, and existing matchers have a hard time finding the correct correspondence.  

Still, the probability of finding a good match when uniformly randomly sampling from $k n_1$ correspondences can be low in practice, leading to many iterations and high runtimes. 
Thus, we follow a PROSAC-like~\cite{chum2005matching} procedure where the potential matches are ordered by a quality prior.
First, we select the correspondence that is the most likely to be correct, and then, progressively, we sample from less likely ones. 
This prior either comes directly from the applied matcher or is predicted by a deep network. 
In this paper, we train the recent  NeFSAC~\cite{cavalli2022nefsac} to predict the probability of each AC leading to an accurate model.
The exact procedure is described in the supp.\ material. 

\vspace{1mm}\noindent\textbf{Scoring and Guided Matching.}
Assume that we are given a model $\theta \in \mathbb{R}^{d_\theta}$ estimated from a single correspondence ($d_\theta \in \mathbb{N}$ is the dimensionality of the model manifold), point sets $\Points_1$ and $\Points_2$ in the two images, and a point-to-model residual function $\phi: \mathbb{R}^{d_\theta} \times \mathbb{R}^{d_p} \to \mathbb{R}$, where $d_p \in \mathbb{N}$ is the data dimension. 
Model $\theta$ can be, for example, an essential matrix and $\phi$ the Sampson distance or symmetric epipolar error. 
In short, we iterate through all potential matches; select the pair with the lowest point-to-model residual for each point in the first image; and, finally, calculate the score from the selected correspondences. 
The pseudo-code for the guided sampling is as follows:
\begin{algorithmic}
\Require $\Points_1$ - points, $\theta$ - model, $H$ - hashing fn.\ \\ $K$ - $k$ best match, $\epsilon$ - thr., $W$ - weight fn., $Q$ - scoring
\Ensure $\mathcal{M}$ - correspondences, $q$ - model score
\State $\mathcal{M} \leftarrow \emptyset$ \Comment{Initialization to empty set}
\ForEach{$\textbf{p}_1 \in \Points_1$} \Comment{Each point in the 1st image}
    \State{$r^* \leftarrow \epsilon$, $\textbf{p}_2^* \leftarrow \textbf{0}$} \Comment{Best residual and match}
    \ForEach{$\textbf{p}_2 \in (K(\textbf{p}_1) \cap H(\textbf{p}_1, \theta))$}
            \If {$\phi((\textbf{p}_1, \textbf{p}_2), \theta) < r^*$}
                \State{$r^* \leftarrow \phi((\textbf{p}_1, \textbf{p}_2), \theta)$, $\textbf{p}_2^* \leftarrow \textbf{p}_2$}
            \EndIf
    \EndFor
    \If {$r^* < \epsilon$}
        \State{$\mathcal{M} \leftarrow \mathcal{M} \cup \{ (\textbf{p}_1, \textbf{p}_2^*) \}$ }
        \State{$q \leftarrow q + W(K(\textbf{p}_1)) Q(\theta)$}
    \EndIf
\EndFor
\end{algorithmic}
The inputs of the algorithm are the points in the first image $\Points_1$; 
model $\theta$; 
a function $K: \Points_1 \to \Points_2^k$ assigning the $k$ best match in the second image to a point in the first one; 
the inlier-outlier threshold $\epsilon \in \mathbb{R}^+$;
a weighting $W: \mathbb{R} \to \mathbb{R}$, 
a model scoring $Q: \mathbb{R}^d \to \mathbb{R}$, 
and a hashing function $H: \Points_1 \times \mathbb{R}^d \to \Points_2^*$.
We use MAGSAC++~\cite{barath2020magsac++} as $Q$ to calculate the model score via marginalizing over an acceptable range of noise scale $\sigma$. 

Given point $\textbf{p}_1$ and model $\theta$, the purpose of the hashing function $H$ is to efficiently select matches from $\Points_2$ that are consistent with $\theta$ when paired $\textbf{p}_1$, \ie, $\forall \textbf{p}_2 \in H(\textbf{p}_1, \theta):   \phi(\textbf{p}_1, \textbf{p}_2) \leq \epsilon$. 
Such $H$ can be easily constructed for homographies or rigid transformations using regular grids. 
Also, one can use epipolar hashing~\cite{barath2021graph} when estimating relative pose. 
In cases, where no such function exists for a particular model, $H$ can be omitted without affecting the accuracy.

We found that it is important to use a weighting $W$ in the score calculation, especially, when estimating relative pose, \ie, fundamental or essential matrix. 
The reason is that the point-to-model residual (\eg, Sampson distance) being zero, does not necessarily mean that it is a correct correspondence.
We are not able to measure the translation along the epipolar lines. 
Without accounting for this, the procedure tends to hallucinate a large amount of incorrect matches that are consistent with the found model. 
The model has lots of inliers, while being incorrect. 
%
Therefore, for cases with such residual functions, we introduce an additional parameter $\mu \in [0, 1]$ that will act similarly as the Lowe ratio threshold~\cite{SIFT2004} or Wald criterion~\cite{Wald1947}. 
For each point $\textbf{p}_1$, we are given $K(\textbf{p}_1) = \{ \textbf{p}_2^1, ..., \textbf{p}_2^k \}$ with matching scores $S(\textbf{p}_1) = \{ s_{12}^1, ... s_{12}^k \}$ from the feature matcher. 
We only keep those potential matches from $K(\textbf{p}_1)$, where the matching score $s_{12}^i \geq \mu \left( \max S(\textbf{p}_1) \right)$. Thus, $K'(\textbf{p}_1) = \{ \textbf{p}_2^i \; | \; \textbf{p}_2^i \in K(\textbf{p}_1) \wedge s_{12}^i \geq \mu \left( \max S(\textbf{p}_1) \right) \}$
Weight $W(\textbf{p}_1) = |K'(\textbf{p}_1)|^{-1}$ in the proposed algorithm.
Thus, the weight is inversely proportional to the number of matches that have similar matching scores. 

\vspace{1mm}\noindent\textbf{Local Optimization.}
As it was discussed in~\cite{barath2020making,barath2022affinetutorial}, inner RANSAC-based local optimization is crucial when using ACs. 
Thus, when a new best model is found, we apply a few iterations of RANSAC on the selected matches using a point-based solver, ignoring the affine shapes. 
For example, this means that the refitting is done by the 5PC~\cite{stewenius2006recent} algorithm when estimating \textbf{E} matrices. 
%
%
In practice, the LO runs only $\log t$ times~\cite{chum2003locally}, where $t$ is the total iteration number of the outer loop. 
The iteration number spent inside the local optimization is typically set to a small value, \eg, $20$. 

\section{Homography from 1AC}

In this section, we propose a new minimal solver for homography estimation using a single affine correspondence as input, and assume the gravity direction to be known. 
While requiring the gravity might seem a restrictive constraint, assuming that it points downwards and is $[0, -1, 0]^\text{T}$ is a reasonably good assumption in practice and it works in all our experiments. 

Homography matrix $\Hom \in \RR^3$ is defined as 
%
    $\Hom = \Rot - \frac{1}{d}\Tran\Normline^{\text{T}}$,
%
where $\Rot \in \text{SO}(3)$ and $\Tran \in \RR^3$ are the relative camera rotation and translation, respectively, $d \in \RR$ is the plane intercept and $\Normline \in \RR^3$ is its normal.
To solve for $\Hom$, 
first, we derive the constraints for relative pose $\Rot, \Tran$ from a single AC $(\Point_1, \Point_2, \Aff)$, and the vertical directions $\Vertical_1 = [x_{v_1}, y_{v_1}, z_{v_1}]^\trans, \Vertical_2 = [x_{v_2}, y_{v_2}, z_{v_2}]^\trans$ known in both images.
The relative pose with a known vertical direction has three degrees-of-freedom (DoF), and the AC imposes three constraints on it.

\begin{figure}
    \centering
    \begin{tabular}{c}
       \includegraphics[width=0.495\linewidth,trim={4mm 0 3mm 0},clip]{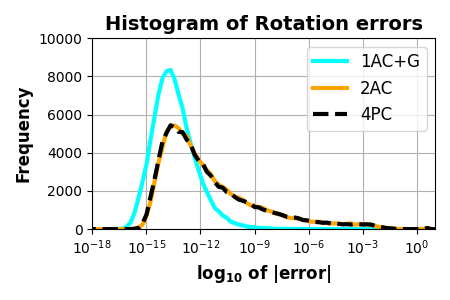} \includegraphics[width=0.495\linewidth,trim={4mm 0 3mm 0},clip]{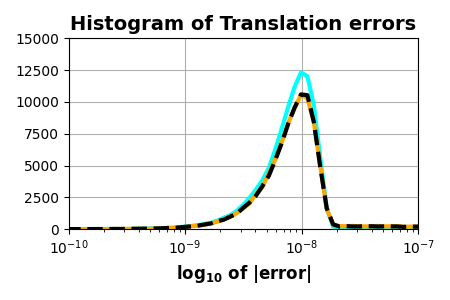} 
    \end{tabular}
    \caption{\textbf{Stability study.} 
    The frequencies ($100$k runs) of $\text{log}_{10}$ rotation and translation errors (both in degrees) in the homography estimated by the 4PC~\cite{hartley2003multiple}, 2AC~\cite{barath2017theory}, and proposed 1AC+G(H) solvers.    }
    \label{fig:stability_tests}
\end{figure}

According to \cite{DBLP:journals/jmiv/KalantariHJG11}, we can express the rotation matrix as $\Rot = \Rot_2^{\text{T}} \Rot_y \Rot_1$, where $\Rot_y$ is a rotation around $y$-axis, $\Rot_1$ transforms $\Vertical_1$ to $y$-axis, $\Rot_2$ transforms $\Vertical_2$ to $y$-axis. Let $\mathbf{y}=[0, \  1, \ 0]^{\text{T}}$ be the $y$-axis. The axis of $\Rot_1$ can be computed as $\Vertical_1 \times \mathbf{y} = [-z_{v_1}/d, \ \ 0, \ \ x_{v_1}/d]^{\text{T}}$, where $d = x_{v_1}^2 + z_{v_1}^2$, the angle is obtained as $\arccos{(\Vertical_1^{\text{T}} \mathbf{y})} = \arccos{(y_{v_1})}$. Rotation matrix $\Rot_1$ is computed using the Rodrigues formula, rotation matrix $\Rot_2$ is obtained similarly. Matrix $\Rot_y$ is expressed elementwise as
\begin{equation}
    \Rot_y = 
    \frac{1}{1+x^2} \begin{bmatrix}
    1-x^2 & 0 & -2x \\ 0 & 1+x^2 & 0 \\ 2x & 0 & 1-x^2 
    \end{bmatrix}, \label{eq:r_y}
\end{equation}
where $x=\tan \phi / 2$. Now, we can express the essential matrix $\Ess$ as $\Ess = \Rot_2^{\text{T}} [\Tran']_\times \Rot_y \Rot_1$, where $\Tran' = \Rot_2 \Tran$.

Let $\Qoint_1 = \Rot_1 \Point_1$ and $\Qoint_2 = \Rot_2 \Point_2$. 
Eq.~\eqref{eq:ess_constraint} becomes
\begin{equation}
    \Qoint_2^{\text{T}} [\Tran']_\times \Rot_y \Qoint_1 = 0. \label{eq:ess_constraint_modified}
\end{equation}
In order to modify constraints \eqref{eq:aff_constraint} in a similar way, we first define $\matr{B} = \Aff^{-\text{T}} [\mathbf{r}_1^1 \, \mathbf{r}_1^2]^{\text{T}}$, $\matr{C} = [\mathbf{r}_2^1 \, \mathbf{r}_2^2]^{\text{T}}$, where $\mathbf{r}_i^1$, $\mathbf{r}_i^2$ $\mathbf{r}_i^3$ are the column vectors of $\Rot_i$, $i \in \{1,2\}$.

The elements of $\matr{B}$ are written in row-major order as $b_1, ..., b_6$, and the elements of $\matr{C}$ as $c_1, ..., c_6$. 
We can rewrite the constraints \eqref{eq:aff_constraint} as
\begin{equation}
\small
\begin{split}
    \Aff^{-\text{T}} \Normline_1 - \Normline_2 = \Aff^{-\text{T}} \Line_{1[1:2]} - \Line_{2[1:2]} \\ 
    = \Aff^{-\text{T}} [\mathbf{r}_1^1 \ \mathbf{r}_1^2 ]^{\text{T}} \Rot_y^{\text{T}} [\Tran']_\times^{\text{T}} \Qoint_2 - [\mathbf{r}_2^1 \ \mathbf{r}_2^2 ]^{\text{T}} [\Tran']_\times \Rot_y \Qoint_1 = 0.
\end{split} \label{eq:aff_constraint_modified}
\end{equation}
Constraints \eqref{eq:ess_constraint_modified}, \eqref{eq:aff_constraint_modified} give $3$ equations in variables $x \in \RR$ and $\Tran' 
\in \RR^3$. 
After multiplying the equations with $1+x^2$, we get three equations that are linear in the elements of translation $\Tran'$.  
We can, therefore, use the \textit{hidden variable approach} to rewrite the equations in the form $\matr{M}(x) \Tran' = 0$, where $\matr{M}(x)$ is a $3 \times 3$ matrix whose elements depend on $x$. 
If $(x, \Tran')$ is a solution to the linear system, then matrix $\matr{M}(x)$ must be singular. Consequently, $\det \matr{M}(x) = 0$ holds. 
This is a univariate polynomial of degree $6$. 
We find its roots as the eigenvalues of its \textit{companion matrix}.
After finding $x$, we calculate $\Tran'$ as the kernel of matrix $\matr{M}(x)$ and the rotation $\Rot_y$ according to \eqref{eq:r_y}. 
Finally, we compute the relative pose $(\Rot, \Tran)$ as $\Rot = \Rot_2^{\text{T}} \Rot_y \Rot_1$, $\Tran = \Rot_2^{\text{T}} \Tran'$.
%

Next, we will solve for the unknown plane parameters using the estimated relative pose. 
We can set $\Normline' = \frac{1}{d}\Normline$ and simplify the expression as follows:
\begin{equation}
    \Hom = \Rot - \Tran\Normline'^{\text{T}}. \label{eq:homography_calibrated}
\end{equation}
To find the homography $\Hom$ consistent with both the affine correspondence $(\Point_1, \Point_2, \Aff)$ and vertical directions $\Vertical_1$ and $\Vertical_2$, we substitute $(\Rot, \Tran)$ into \eqref{eq:homography_calibrated}. 
Then, we only need to find vector $\Normline' \in \RR^3$. 
We substitute the expression \eqref{eq:homography_calibrated} into the constraints from~\cite{DBLP:journals/tip/BarathH18} connecting affine correspondences and homography $\Hom$.
We obtain $6$ linear equations in $3$ unknowns.
They are shown in the supp.\ mat.
%
%
The LS method obtains vector $\Normline'$ from the above system. 
Finally, we compute the homography $\Hom$ from $\Rot$, $\Tran$, $\Normline'$ using the equation \eqref{eq:homography_calibrated}.

\section{Experiments}

This section first tests the proposed minimal solver in a fully controlled synthetic environment.
Then \textit{AffineGlue} is evaluated on real-world datasets for relative pose and homography estimation. 
All experiments were implemented in C++ and performed on an Intel(R) Core(TM) i9-10900K CPU @ 3.70GHz.

\begin{figure}
    \centering
    \setlength\tabcolsep{0pt}
    \setlength\extrarowheight{-3pt}
    \renewcommand{\arraystretch}{0}
    \begin{tabular}{c c}
       \includegraphics[width=0.5\linewidth]{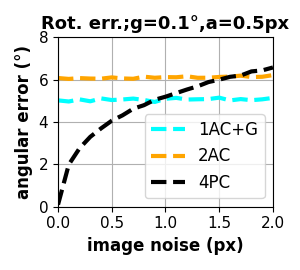}
       &
       \includegraphics[width=0.5\linewidth]{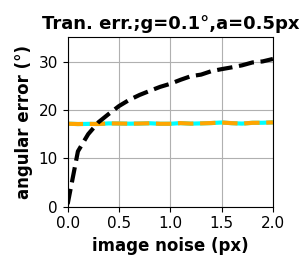}
       \\

%
%
    \end{tabular}
    \caption{\textbf{Image noise study.}
    The average (over $100$k runs) angular errors of the rotations and translation estimated by the 4PC~\cite{hartley2003multiple}, 2AC~\cite{barath2017theory}, and proposed 1AC+G(H) homography solvers plotted as a function of the image noise in pixels.}
    \label{fig:noise_tests_H}
\end{figure}

\vspace{1mm}\noindent\textbf{Synthetic Experiments.}
%
To create a synthetic scene, we generate two cameras with random rotations and translations and focal length set to $1000$.
A randomly oriented 3D point is generated and projected into both cameras.
The affine transformation is calculated from the point orientation. 
We generated $100$k random problem instances and ran the solvers on noiseless samples. 
Fig.~\ref{fig:stability_tests} shows histograms of the $\log_{10}$ rotation and translation errors. 
The plots show that all solvers are stable -- there is no peak close to $10^0$.
In Fig.~\ref{fig:noise_tests_H}, the average errors in degrees are shown as a function of the image noise. 
%
We use a fixed gravity (0.1$^{\circ}$) and affine noise ($0.5$ px).
It is important to note that the realistic affine noise is unclear in practice, with no work analyzing it. 
These plots only intend to demonstrate that the solvers act reasonably w.r.t.\ increasing noise levels, which they do. 
%
More synthetic experiments are in the supplementary material.

%
%
%
%

\begin{table*}[]
    \centering
    \resizebox{1.0\textwidth}{!}{\begin{tabular}{r c r | c c c c c c c c}
         \multicolumn{1}{c}{Features} & Estimator & Solver & AVG $\downarrow$ & MED $\downarrow$ & AUC@$1^\circ \uparrow$ & @$2.5^\circ \uparrow$ & @$5^\circ \uparrow$ & @$10^\circ \uparrow$ & @$20^\circ \uparrow$ & \# inliers \\
         \midrule
         \midrule
         \multirow{4}{*}{SuperPoint + SuperGlue} & \multirow{2}{*}{AffineGlue} & 1AC+$i$G &  \phantom{1}\abest{2.6} & \abest{0.7} & \textbf{34.5} & {55.9} & {70.3} & {81.3} & \abest{89.2} & {394} \\
         &  & 1AC+\textit{m}D & \phantom{1}\abest{2.6} & 0.8 & \textbf{34.5} & \textbf{56.0} & \abest{70.4} & \abest{81.4} & \abest{89.2} & 395 \\
          & \multirow{2}{*}{MAGSAC++} & 5PC & \phantom{1}4.1 & 1.3 & 23.0 & 43.5 & 59.9 & 74.1 & 84.6 & 276 \\
         & & 1PC+$i$G & \phantom{1}4.0 & 1.3 & 23.0 & 43.4 & 59.6 & 74.0 & 84.7 & 276 \\
         \midrule
         
         \multirow{4}{*}{DoG-8k + HardNet + AffNet} & \multirow{2}{*}{AffineGlue} & 1AC+$i$G & \phantom{1}\textbf{{3.4}} & \abest{0.7} & \abest{38.7} & \abest{57.4} & \textbf{70.0} & \textbf{79.9} & \textbf{87.4} & {286} \\
          & & 1AC+\textit{m}D & \phantom{1}5.2 & 0.9 & 22.2 & 50.6 & 62.6 & 73.0 & 81.7 & 202  \\
         & \multirow{2}{*}{MAGSAC++} & 5PC  & \phantom{1}6.3 & 1.4 & 27.7 & 42.7 & 54.3 & 66.2 & 77.2 & 210 \\
         & & 1AC+$i$G & \phantom{1}5.1 & 0.9 & 33.3 & 50.5 & 62.5 & 72.9 & 81.6 & 257 \\
         \midrule
         DoG-8k + HardNet + Adalam & \multirow{9}{*}{\rotatebox{90}{\small MAGSAC++}} & 5PC & \phantom{1}8.8 & \textbf{0.8} & \textbf{34.3} & \textbf{52.5} & \textbf{65.0} & \textbf{74.8} & 82.4 & 307 \\  
         {LoFTR} & & 5PC & \phantom{1}\textbf{3.6} & 1.3 & 22.5 & 43.4 & 59.6 & 73.7 & \textbf{84.5} & 866 \\
          LoFTR & & 3PC+$i$G & \phantom{1}4.1 & 1.4 & 21.0 & 40.9 & 56.7 & 71.1 & 82.6 & 878 \\
         {DISK} & & 5PC & \phantom{1}4.7 & 0.9 & 27.9 & 44.3 & 55.7 & 64.5 & 71.2 & 474 \\
          DISK & & 3PC+$i$G & \phantom{1}4.5 & \textbf{0.8} & 29.1 & 45.8 & 57.1 & 66.1 & 72.9 & 617 \\
         {R2D2 + NN} & & 5PC & 13.0 & 2.7 & 13.6 & 28.8 & 42.9 & 57.9 & 70.3 & 169 \\
          R2D2 + NN & & 3PC+$i$G & 12.9 & 2.7 & 13.9 & 28.8 & 42.8 & 57.5 & 70.2 & 169 \\ 
         {DoG-8k + SOSNet + NN} & & 5PC & 40.4 & 5.9 & 12.8 & 23.9 & 33.5 & 43.3 & 52.9 & \phantom{1}55 \\
         DoG-8k + SOSNet + NN & & 3PC+$i$G & 40.4 & 5.9 & 12.9 & 23.8 & 33.4 & 43.3 & 52.9 & \phantom{1}55 \\
    \end{tabular}}
    \caption{
    \textbf{Relative pose estimation on PhotoTourism}~\cite{IMC2020} on a total of 9900 image pairs.
    We report the avg.\ and median pose errors (in degrees; max.\ of the translation and rotation errors), their AUC scores and the inlier numbers. 
    We use the 3PC+\textit{i}G~\cite{ding2020efficient} and the 1AC+\textit{i}G~\cite{guan2019rotational}  solvers with identity gravity, the 1AC+\textit{m}D solver~\cite{eichhardt2020relative} on depth from MiDaS-v3~\cite{Ranftl2021,Ranftl2022}, and
    the five point method (5PC)~\cite{Nister2004}. 
    For solvers requiring more than a single match, we apply the state-of-the-art MAGSAC++~\cite{barath2020magsac++}. 
    Finally, the Levenberg-Marquardt method~\cite{more1978levenberg} minimizes the pose error on all inliers.
    The best values are bold in each group. 
    The absolute best ones are underlined.}
    \label{tab:results_phototourism}
\end{table*}

\subsection{Real-World Experiments}

\begin{table*}[]
    \centering
    \resizebox{1.0\textwidth}{!}{\begin{tabular}{r c r | c c c c c c c c}
         \multicolumn{1}{c}{Features} & Estimator & Solver & AVG $\downarrow$ & MED $\downarrow$ & AUC@$1^\circ \uparrow$ & @$2.5^\circ \uparrow$ & @$5^\circ \uparrow$ & @$10^\circ \uparrow$ & @$20^\circ \uparrow$ & \# inliers \\
         \hline
         \hline
         \multirow{4}{*}{SuperPoint + SuperGlue} & \multirow{2}{*}{AffineGlue} & 1AC+$i$G & \abest{12.9} & \phantom{1}{5.8} & {\textbf{0.8}} & {\textbf{7.1}} & {{20.6}} & \textbf{39.7} & \abest{58.4} & 119 \\
         & & 1AC+\textit{m}D & {14.0} & \phantom{1}\abest{5.5} & \textbf{0.8} & 7.0 & {\textbf{20.7}} & {39.8} & {58.1} & {110} \\
         & \multirow{2}{*}{MAGSAC++} & 5PC & 21.4 & \phantom{1}6.5 & 0.7 & 5.9 & 17.3 & 33.9 & 50.9 & \phantom{1}89 \\
         & & 3PC+\textit{i}G & 32.4 & 21.0 & 0.5 & 4.2 & 11.5 & 21.9 & 33.1 & \phantom{1}84 \\
         \midrule
         \multirow{4}{*}{DoG-8k + HardNet + AffNet} & \multirow{2}{*}{AffineGlue} & 1AC+$i$G & 26.8 & 15.0 & \textbf{0.7} & \textbf{5.0} & \textbf{13.0} & 24.2 & 37.2 & 146 \\
          & & 1AC+\textit{m}D & {\textbf{24.7}} & {\textbf{12.4}} & 0.6 & 4.5 & 12.6 & \textbf{25.3} & \textbf{39.6} & 120 \\
          & \multirow{2}{*}{MAGSAC++} & 5PC & 33.7 & 29.9 & 0.3 & 2.3 & \phantom{1}6.6 & 13.6 & 22.9 & \phantom{1}81 \\
          & & 1AC+\textit{i}G & 25.3 & 13.0 & 0.3 & 3.1 & \phantom{1}9.0	& 18.4 & 29.4 & \phantom{1}64 \\
         \midrule
         DoG-8k + HardNet + Adalam & \multirow{6}{*}{\rotatebox{90}{MAGSAC++}} & 5PC & 54.1 & 17.8 & 0.5 & 3.7 & 11.1 & 22.3 & 34.9 & 101
         \\
         %
         LoFTR & & 5PC & {30.3} & \phantom{1}\textbf{6.6} & \abest{1.1} & \abest{8.3} & \abest{22.5} & \abest{41.2} & \textbf{57.7} & 468 \\
         	
         R2D2 + NN & & 5PC & 32.9 & 13.6 & 0.6 & 4.2 & 12.0 & 24.6 & 38.1 & 190 \\
         R2D2 + NN & & 3PC+\textit{i}G & \textbf{18.9} & 10.6 & 0.4 & 2.8 & \phantom{1}8.2 & 16.8 & 27.4 & 137\\
         DoG-8k + SOSNet + NN & & 5PC & 33.3 & 29.7 & 0.4 & 2.6 & \phantom{1}6.6 & 13.6 & 23.4 & \phantom{1}78 \\
         DoG-8k + SOSNet + NN & & 3PC+\textit{i}G & 60.8 & 36.4 & 0.3 & 1.6 & \phantom{1}5.3 & 12.4 & 22.5 & \phantom{1}38 \\
    \end{tabular}}
    \caption{
    \textbf{Relative pose estimation on ScanNet}~\cite{dai2017scannet} on the 1500 image pairs from~\cite{sarlin2020superglue,sun2021loftr}.
    We report the avg.\ and median pose errors (in degrees; max.\ of the translation and rotation errors), their AUC scores and the inlier numbers. 
    We use the 3PC+\textit{i}G~\cite{ding2020efficient} and 1AC+\textit{i}G~\cite{guan2019rotational} solvers with identity gravity, the 1AC+\textit{m}D solver~\cite{eichhardt2020relative} on depth from MiDaS-v3~\cite{Ranftl2021,Ranftl2022}, and
    the five point method (5PC)~\cite{Nister2004}. 
    For solvers requiring more than a single match, we apply the state-of-the-art MAGSAC++~\cite{barath2020magsac++}. 
    Finally, the Levenberg-Marquardt method~\cite{more1978levenberg} minimizes the pose error on all inliers.
    The best values are bold in each group. 
    The absolute best ones are underlined.}
    \label{tab:results_scannet}
\end{table*}

\begin{table*}[]
    \centering
    \resizebox{0.9\textwidth}{!}{\begin{tabular}{r c r | c c c c c }
         \multicolumn{1}{c}{Features} & Estimator & Solver & AUC@$1\text{px} \uparrow$ & @$2.5\text{px} \uparrow$ & @$5\text{px} \uparrow$ & @$10\text{px} \uparrow$ & Time (secs) $\downarrow$ \\ 
         \hline
         \hline
         \multirow{3}{*}{SuperPoint + SuperGlue} & AffineGlue & 1AC+$i$G-H & \abest{50.5} & \abest{73.9} & \abest{84.9} & \abest{91.1} & \textbf{0.04} \\ 
          & \multirow{2}{*}{{MAGSAC++}} & 1AC+$i$G-H & 45.6 & 71.7 & 83.9 & 90.9 & 0.66 \\ 
          &  & 4PC & 37.9 & 65.6 & 79.0 & 90.1 & 0.60 \\ 
         \midrule
         \multirow{3}{*}{DoG-2k + HardNet + AffNet} & AffineGlue & 1AC+$i$G-H & 40.1 & 68.0 & 81.4 &  88.8 & 0.29 \\ 
          & \multirow{2}{*}{{MAGSAC++}} & 1AC+$i$G-H & 40.3 & 68.8 & 82.3 & 89.8 & 0.11  \\ 
          &  & 4PC & \textbf{40.9} & \textbf{69.3} & \textbf{82.7} & \textbf{90.4} & \abest{0.01} \\ 
         \midrule
         LoFTR & \multirow{7}{*}{\rotatebox{90}{MAGSAC++}} & 4PC & \textbf{41.8} & \textbf{68.6} & \textbf{81.2} & \textbf{87.9} & 0.40 \\
         DoG-2k + SOSNet + NN &  & 1AC+$i$G-H & {38.3}	& {65.5} & {79.5} & {87.4} & 0.47 \\ 
         DoG-2k + SOSNet + NN &  & 4PC & 36.9 & 63.3 & 77.0 & 85.1 & 0.25 \\ 
         R2D2 + NN &  & 1AC+$i$G-H & 27.6 & 51.5 & 65.9 & 75.1 & 0.20 \\ 
         R2D2 + NN & & 4PC  & 27.4 & 51.0 & 65.5 & 75.4 & \textbf{0.09} \\ 
         DISK + NN &  & 1AC+$i$G-H & 25.1 & 51.8 & 68.5 & 77.8 & 0.29 \\ 
         DISK + NN & & 4PC & 25.0 & 51.5 & 68.1 & 78.7 & 0.20 \\ 
    \end{tabular}}
    \caption{
    \textbf{Homography estimation} on the HPatches dataset~\cite{balntas2017hpatches}.
    The AUC scores and avg.\ times are reported. 
    AffineGlue is applied with the proposed 1AC+\textit{i}G-H solver assuming identity gravity. 
    We also run MAGSAC++~\cite{barath2020magsac++} with the 4PC~\cite{hartley2003multiple} and 1AC+\textit{i}G-H solvers.
    The best values are bold in each group, the absolute bests are underlined.}
    \label{tab:results_hpatches}
\end{table*}

\noindent\textbf{Affine Features.}
There are multiple ways to obtain affine features from real images. 
First, the most standard is to use a local feature detector, like DoG~\cite{SIFT2004} or Key.Net~\cite{KeyNet2019}, estimate keypoint locations and scales, and use the patch-based AffNet~\cite{AffNet2018} to get affine shapes. 
Finally, a patch-based descriptor, like HardNet~\cite{HardNet2017} or SOSNet~\cite{SoSNet2019}, is applied. 
This approach is among leaders in IMC 2020 benchmark~\cite{IMC2020}.

The second way is to use handcrafted affine detectors, such as MSER~\cite{MSER2002} and W$\alpha$SH~\cite{wash2012}, that jointly estimate local feature geometry including affine shape. 
On top of these features, we can detect any patch-based descriptors, \eg, HardNet~\cite{HardNet2017} or SOSNet~\cite{SoSNet2019}.

Finally, we experimented with joint detector-descriptor models, such as SuperPoint~\cite{SuperPoint2017} and DISK~\cite{DISK2020}, which outputs keypoint location and descriptor. 
To upgrade point-features to affine-features, we employ Self-Scale-Ori~\cite{selfscaori2021} scale estimator to get the scale and orientation. 
Finally, AffNet runs to get affine shape. 
Note, it gives a user 2 options -- either use original SuperPoint/DISK descriptors or patch-based HardNet on top of affine feature. 

In the main experiments, we run the proposed \textit{AffineGlue} on DoG + HardNet + AffNet + NN (NN -- nearest neighbor matching) and SuperPoint + Self-Scale-Ori + AffNet + SuperGlue features since they lead to the most accurate results -- this will be shown in the ablation study. 
Obtaining a pool of potential matches is straightforward when using NN on HardNet descriptors. 
To get a similar pool for SuperGlue, we directly access the matching score matrix that is obtained when solving the optimal transport problem. This allows selecting the $k$ best matches for each point. 

\vspace{1mm}\noindent\textbf{Minimal Solvers.}
When testing relative pose estimation, we compare three solvers.
5PC~\cite{stewenius2006recent} is the widely-used algorithm estimating the pose from five point correspondences. 
The 1AC+\textit{m}D solver is proposed in~\cite{eichhardt2020relative}.
It estimates the pose from a single AC and predicted monocular depth.
To allow running this solver, we obtain relative depth by MiDaS-v3~\cite{Ranftl2021,Ranftl2022}.
We also compare solver 1AC+G~\cite{guan2019rotational} that requires a single AC and a known direction in the images. 
To demonstrate the robustness of the proposed \textit{AffineGlue}, we \textit{always} run 1AC+G assuming that the gravity points downwards -- its direction is $[0, -1, 0]^\text{T}$. 
Thus, we call the solver 1AC+$i$G. 
This way, we do not require to know the gravity direction prior to running the algorithm.
This is based on two assumptions that proved true on the tested datasets: (i) people tend to roughly align their cameras with the gravity direction~\cite{Perdoch-CVPR2009efficient,IMC2020}; (ii) \textit{AffineGlue} is robust enough so if the estimated noisy model is able to select a few inliers, the local optimization procedure recovers. 
We also test the 3PC+G~\cite{ding2020efficient} solver that requires three PCs and the gravity. Similarly as before, we use identity gravity. 

\vspace{1mm}\noindent\textbf{Relative Pose -- PhotoTourism.} 
For testing the methods, we use the data from the CVPR IMC 2020 PhotoTourism challenge~\cite{IMC2020}. 
It consists of 25 scenes (2 -- validation; 12 -- training; 11 -- test sets) of landmarks with photos of varying sizes collected from the internet. 
NeFSAC is trained by splitting the training set into two disjoint sets for training and validation.
The algorithms are tested on the two scenes for validation -- a total of 9900 pairs.
For robust estimation, we chose MAGSAC++~\cite{barath2020magsac++} as competitor. 
We compare the following detectors: SuperPoint~\cite{SuperPoint2017} with SuperGlue~\cite{sarlin2020superglue}, DoG~\cite{SIFT2004} with HardNet~\cite{HardNet2017} descriptors, DoG with HardNet followed by Adalam~\cite{cavalli2020handcrafted}, DoG with SOSNet~\cite{SoSNet2019} descriptors,  DISK~\cite{DISK2020}, and R2D2~\cite{R2D22019}. 
Also, we show the results of LoFTR~\cite{sun2021loftr}.
The average error of the gravity prior $[0, -1, 0]^\text{T}$ is $10.8^\circ$.

The results are in Table~\ref{tab:results_phototourism}.
We report the average and median pose errors (\ie, the max.\ of the rotation and translation errors) in degrees, the AUC scores at $1^\circ$, $2.5^\circ$, $5^\circ$, $10^\circ$, and $20^\circ$, and the average inlier number.
Note that the inlier number is not informative when different detectors and matchers are compared. 
We show it to highlight that the proposed method increases the inlier number compared to MAGSAC++ with 5PC on the same features.
DoG+HardNet+\textit{AffineGlue} and SP+SG+\textit{AffineGlue}, on par, lead to the best results. 
Compared to the best method with MAGSAC++ (\ie, DoG+HardNet+Adalam+5PC), DoG+HardNet+\textit{AffineGlue} improves at least 5 AUC points in \textit{all} metrics. 
Moreover, let us highlight that using the AC+$i$G solver instead of 5PC in MAGSAC++, improves DoG+HardNet by a large margin, \ie, $5$-$8$ AUC points. 
Interestingly, using the 3PC+$i$G~\cite{ding2020efficient} solver only marginally improves the results of MAGSAC++. 
There is no significant difference in the results of the 1AC+$i$G and 1AC+$m$D solvers.
Thus, we suggest using the 1AC+$i$G as it does not require running a depth predictor.

\vspace{1mm}\noindent\textbf{Relative Pose -- ScanNet.} 
The ScanNet dataset~\cite{dai2017scannet} contains 1613 monocular sequences with ground truth poses and depth maps.
We evaluate our method on the 1500 pairs used in SuperGlue~\cite{sarlin2020superglue} and \cite{sun2021loftr}.
These pairs contain wide baselines and extensive texture-less regions.
The avg.\ error of the gravity prior is $24.8^\circ$.

The results are shown in Table~\ref{tab:results_scannet}. 
We can see similar results as for PhotoTourism. 
\textit{AffineGlue} with DoG or SuperPoint+SuperGlue features improves the performance by a large margin.  
It makes SuperPoint+SuperGlue comparable to the detector-less LoFTR~\cite{sun2021loftr} with achieving even smaller avg.\ and med.\ errors and higher AUC@$20^\circ$.
DoG+HardNet with \textit{AffineGlue} is less accurate than SP+SG, however, it still is among the top-performing methods. 
Both 1AC+\textit{i}G and 1AC+\textit{m}D lead to similar accuracy.

\vspace{1mm}\noindent\textbf{Homography -- HPatches.}
The~\cite{balntas2017hpatches} dataset contains 52 sequences under significant illumination changes and 56 sequences that exhibit large viewpoint variation.
Since the intrinsic matrices are not provided in HPatches, we calibrate the cameras of the 56 sequences with viewpoint changes by the RealityCapture software~\cite{realitycapture}.
We use these sequences in the evaluation. 

The results are shown in Table~\ref{tab:results_hpatches}.
The proposed \textit{AffineGlue} with SuperPoint+SuperGlue leads to the most accurate results while being one of the fastest algorithms.
Its AUC@1$^\circ$ score is increased by $5$ AUC points compared to the second most accurate method. 

\vspace{1mm}\noindent\textbf{Run-time.}
As reported in Table~\ref{tab:results_hpatches}, the avg.\ run-time of \textit{AffineGlue} on \textbf{H} estimation from SuperPoint+SuperGlue features is $0.04$ seconds. 
The avg.\ time of pose estimation on PhotoTourism is $0.09$ and on ScanNet is $0.03$ seconds. 
The avg.\ inference time of NeFSAC is $1.1$ ms per image pair.
For comparison, MAGSAC++ with the 5PC solver runs, on average, for $0.01$ secs on ScanNet and for $0.04$ secs on PhotoTourism.
Even though \textit{AffineGlue} is slower, it still runs in real-time while achieving state-of-the-art accuracy.

\vspace{1mm}\noindent\textbf{Feature Ablation.}
We compared a number of affine detectors to choose the best ones.
The AUC scores on PhotoTourism are shown in Table~\ref{tab:feature_ablation_phototourism} and on ScanNet in Table~\ref{tab:feature_ablation_scannet}.
On PhotoTourism, we used the 1AC+\textit{i}G solver.
On ScanNet, we used 1AC+\textit{m}D.
All methods use \textit{AffineGlue}. 
DoG with HardNet and AffNet is on par with SuperPoint with SuperGlue on PhotoTourism.
On ScanNet, SP+SG is the best.
%
Interestingly, SuperPoint works better with HardNet descriptors than its own when NN matching is used.
As expected, classical affine shape detectors, \ie MSER and W$\alpha$SH, are inaccurate even with HardNet descriptors. 


\begin{table}[]
    \centering
    \resizebox{1.0\columnwidth}{!}{\begin{tabular}{r c c | c c c c c}
         Detector & Desc.\ & +AffNet &  AUC@$1^\circ$ & $2.5^\circ$ & $5^\circ$ & $10^\circ$ & $20^\circ$ \\
         \hline
         DoG-8k~\cite{SIFT2004} & \multirow{6}{*}{\rotatebox{90}{HardNet+NN}} & \cmark & 0.5 & 4.5 & 12.6 & 25.3 & 39.6  \\
         SP~\cite{SuperPoint2017} & & \cmark & 0.4 & 2.6 & \phantom{1}7.7 & 16.3 & 26.9 \\
         DISK~\cite{DISK2020} & & \cmark & 0.3 & 2.2 & \phantom{1}6.3 & 13.4 & 21.3 \\
         Key.Net~\cite{KeyNet2019} & & \cmark & 0.3 & 1.8 & \phantom{1}5.3 & 10.7 & 17.4 \\
         MSER~\cite{MSER2002} & & \xmark & 0.1 & 1.2 & \phantom{1}3.5 & \phantom{1}7.2 & 12.5 \\
         W$\alpha$SH~\cite{wash2012} & & \xmark & 0.0	& 0.1 & \phantom{1}0.5 & \phantom{1}1.9 & \phantom{1}5.7 \\
         SP~\cite{SuperPoint2017} & +NN & \cmark & {0.6} & 4.2 & 11.7 & 23.1 & 36.1 \\
         SP~\cite{SuperPoint2017} & +SG & \cmark & \textbf{0.8} & \textbf{7.0} & \textbf{20.7} & \textbf{39.8} & \textbf{58.1}  \\
         DISK~\cite{DISK2020} & +NN & \cmark & 0.3 & 2.4 & \phantom{1}7.2 & 14.7 & 25.1	 \\
    \end{tabular}}
    \caption{\textbf{Affine features on Scannet}~\cite{dai2017scannet} used inside \textit{AffineGlue} on a total of 1500 image pairs.}
    \label{tab:feature_ablation_scannet}
\end{table}

\begin{table}[]
    \centering
    \resizebox{1.0\columnwidth}{!}{\begin{tabular}{r c c | c c c c c}
         Detector & Desc.\ & +AffNet & AUC@$1^\circ$ & $2.5^\circ$ & $5^\circ$ & $10^\circ$ & $20^\circ$ \\
         \hline
         DoG-8k~\cite{SIFT2004} & \multirow{6}{*}{\rotatebox{90}{HardNet+NN}} & \cmark & \textbf{38.7} & \textbf{57.4} & {70.0} & {79.9} & {87.4} \\
         Key.Net~\cite{KeyNet2019} & & \cmark & 22.6 & 38.8 & 51.1 & 62.7 & 73.6 \\
         DISK~\cite{DISK2020} & & \cmark & 16.4 & 27.7 & 37.9 & 49.6 & 63.0 \\
         MSER~\cite{MSER2002} & & \xmark & 13.6 & 24.3 & 34.4 & 46.2 & 58.6 \\
         SP~\cite{SuperPoint2017} & & \cmark & 11.5 & 22.0 & 31.6 & 42.9 & 55.4 \\
         W$\alpha$SH~\cite{wash2012} & & \xmark & \phantom{1}0.0 & \phantom{1}0.1 & \phantom{1}0.8 & \phantom{1}4.0 & 13.6 \\
         SP~\cite{SuperPoint2017} & +NN & \cmark & \phantom{1}8.7 & 17.5 & 26.4 & 37.0 & 48.7 \\
         SP~\cite{SuperPoint2017} & +SG & \cmark & 34.5 & 55.9 & \textbf{70.3} & \textbf{81.3} & \textbf{89.2} \\
         DISK~\cite{DISK2020} & +NN & \cmark & 30.1 & 47.3 & 59.5 & 69.6 & 77.7 \\
    \end{tabular}}
    \caption{\textbf{Affine features on PhotoTourism}~\cite{IMC2020} used inside \textit{AffineGlue} on a total of 9900 image pairs.}
    \label{tab:feature_ablation_phototourism}
\end{table}

\section{Conclusion}

We propose \textit{AffineGlue} to jointly perform feature matching and robust estimation by leveraging a pool of one-to-many correspondences.
Thus, it is significantly less sensitive to matching ambiguities than using traditional top-1 matches.
\textit{AffineGlue} significantly improves performance when applied on top of popular feature detectors and matchers, such as SIFT or SuperPoint+SuperGlue.
Although the used solvers assume that the gravity direction is known in both images, \textit{AffineGlue} is so robust that the $[0, -1, 0]^\text{T}$ gravity prior works even on ScanNet, where it is only a rough approximation with an avg.\ error of $24.8^\circ$ compared to the actual vertical direction.

\vspace{1mm}\noindent\textbf{Limitations and Future Directions.}
One limitation is that most detectors and matchers do not consider feature scale, orientation, and affine shape.
The only exception is the DoG + AffNet combination, where AffNet was trained on DoG detections.
We believe that creating end-to-end affine-covariant features could boost the performance of an \textit{AffineGlue}-based approach.
Additionally, considering the AC in the matching procedure could further improve accuracy, \eg, by training SuperGlue on affine-aware descriptors.


{\small
\bibliographystyle{ieee_fullname}
\bibliography{egbib}
}

\end{document}